\documentclass[letterpaper,10pt, conference]{_Aux/ieeeconf}

\IEEEoverridecommandlockouts
\usepackage{times}
\usepackage{setspace}
\usepackage{url}
\spacing{1}
\usepackage[utf8]{inputenc}
\usepackage[T1]{fontenc}
\usepackage{graphicx}		
\usepackage{wrapfig}
\usepackage[format=plain,font=footnotesize,labelfont=bf,labelsep=period]{caption}
\usepackage{sidecap} 
\usepackage[export]{adjustbox}
\usepackage{subcaption}
\usepackage[font=small]{caption}
\usepackage{float}

\usepackage{comment}

\usepackage{amsmath} 
\usepackage{amssymb}  
\usepackage{amsthm}
\usepackage{mathtools}
\usepackage[normalem]{ulem}
\usepackage{paralist}	
\usepackage[space]{grffile} 
\usepackage{color}
\let\labelindent\relax
\usepackage{enumitem}
\usepackage{bm}
\usepackage{cancel}
\usepackage{hhline}
\usepackage[c2 , nocomma]{optidef}
\usepackage{oubraces}
\usepackage{algorithmic}
\usepackage{graphicx}
\usepackage{textcomp}
\usepackage{siunitx}

\usepackage{moreverb, url}

\usepackage[ruled,vlined]{algorithm2e}
\usepackage{stackengine}
\usepackage{dirtytalk}

\usepackage{array}
\usepackage{verbatim}
\usepackage{upgreek}

\usepackage{amsfonts}
\usepackage{lipsum}
\usepackage{cite}
\usepackage{hyperref}

\theoremstyle{definition}

\theoremstyle{remark}

\theoremstyle{definition}
\theoremstyle{definition}

\definecolor{blue}{RGB}{38,38,134}
\definecolor{darkblue}{RGB}{0,0,102}
\definecolor{lightblue}{RGB}{77,77,148}

\definecolor{gold}{RGB}{234, 170, 0}
\definecolor{metallic_gold}{RGB}{139, 111, 78}

\newcommand{\bs}[1]{\boldsymbol{ #1 }}

\usepackage{xfrac}

\usepackage{dblfloatfix}

\usepackage{pdfpages}
\usepackage{graphicx}

\usepackage{enumitem}

\newcommand{\huseyinrev}[1]{\textcolor[rgb]{0.0, 0.0, 0.0}{#1}}

\newcommand{\johannesrev}[1]{\textcolor[rgb]{0.0, 0.0, 0.0}{#1}}

\begin{document}

\title{ \bf
Earth to Moon Bilateral Teleoperation for Lunar Assembly: A Unified Framework for Shared Autonomy, Skill Learning, and Telepresence

\author{Huseyin Tugcan Dinc$^{1}$, Johannes Heidersberger$^{2}$, Christos Kyriazis$^{3}$, Tobias Wagner$^{1}$, Alexandre Frantz$^{3}$, \\
Dario Pisanti$^{3}$, Fanyi Kong$^{1}$, Miguel A. Olivares-Mendez$^{3}$, Dongheui Lee$^{2,4}$, Christian Ott$^{1,4}$
\thanks{This work was supported in part by the Austrian Science Fund (FWF) under the project ``LunarAssembly'' [10.55776/PIN5145724] and in part by the Luxembourg National Research Fund (FNR) under the LunarAssembly project [18990339].}
\thanks{$^{1}$Huseyin Tugcan Dinc, Tobias Wagner, Fanyi Kong, and Christian Ott are with the Automation and Control Institute, TU Wien, Vienna, Austria.
        {\tt\footnotesize \{hueseyin.dinc@tuwien.ac.at\}}}
\thanks{$^{2}$Johannes Heidersberger and Dongheui Lee are with Autonomous Systems Lab, Institute of Computer Technology, TU Wien, Vienna, Austria.
{\tt\footnotesize \{johannes.heidersberger@tuwien.ac.at\}}}
\thanks{$^{3}$Christos Kyriazis, Alexandre Frantz, Dario Pisanti, and Miguel A. Olivares-Mendez are with Space Robotics Research Group (SpaceR), Interdisciplinary Centre for Security, Reliability and Trust (SnT), University of Luxembourg, Luxembourg.
        {\tt\footnotesize \{christos.kyriazis@uni.lu\}}}
\thanks{$^{4}$Dongheui Lee and Christian Ott are also with Institute of Robotics and Mechatronics, German Aerospace Center (DLR), Weßling, Germany.}}
}

\maketitle 

\begin{abstract}
\huseyinrev{Assembling a lunar base from Earth requires operation across a round-trip delay of around three seconds, making direct teleoperation demanding, while autonomy alone cannot handle all situations remotely. We present a unified Earth-to-Moon robotic manipulation framework that combines learning from demonstration, shared autonomy, and  delayed direct teleoperation with passivity guaranty. We teach specific assembly skills in a delay-free lunar analog test environment using learning from demonstration and transfer the learned skills to the remote site through task parametrization, where the execution can be corrected online through shared autonomy. In the event of failure during autonomous task execution, we provide passivity guaranteed direct teleoperation, enabling the operator to correct and finalize the task.  During direct teleoperation, to increase the operator’s awareness, we use capability maps to help the operator gauge the distance to the target object. Extended reality interfaces further enhance situational awareness through environment reconstruction and visualization of terrain and obstacles. 
}

\end{abstract}

\section{Introduction}
Bilateral teleoperation allows a human operator to remotely control a robot and, at the same time, feel the forces arising from the robot's interaction with its environment. Using this force feedback, the operator does not only send commands to the robot but also senses the remote environment through the interaction forces measured at the robot side \cite{hokayem2006bilateral}. This capability is especially useful for space applications, such as lunar surface operations or in-orbit servicing, where sending a human operator is costly, dangerous, and in some cases may not be feasible, but tasks such as assembly and maintenance still benefit from direct human control and judgment. Teleoperated assembly tasks are especially relevant for building and maintaining permanent infrastructure at the Lunar South Pole (LSP), identified as a priority site for in-situ resource utilization, as underscored by NASA's recent \textit{Ignition} initiative \cite{nasa2026ignition}. 

A major challenge for bilateral teleoperation in space is communication delay. Delay breaks the passivity of the force feedback loop, which causes oscillations that disturb the operator and can damage the remote environment \cite{panzirsch2020safe, dinc2024relaxing}. For lunar teleoperation, the round-trip communication delay ranges from a physical lower bound of $\sim$3 seconds up to $\sim$10 seconds, as in the communication architecture adopted for the VIPER mission \cite{fong2026viper}. Prior work has shown that as round-trip delay increases, operators increasingly rely on a move-and-wait strategy and direct control becomes progressively more difficult due to rising cognitive load \cite{kim2021impact}. \huseyinrev{A recent work demonstrated force-feedback teleoperation without requiring a move-and-wait strategy under sub-second communication delays~\cite{panzirsch2022exploring}.} 

\begin{figure}[t]
    \centering
    \includegraphics[width=0.9\linewidth]{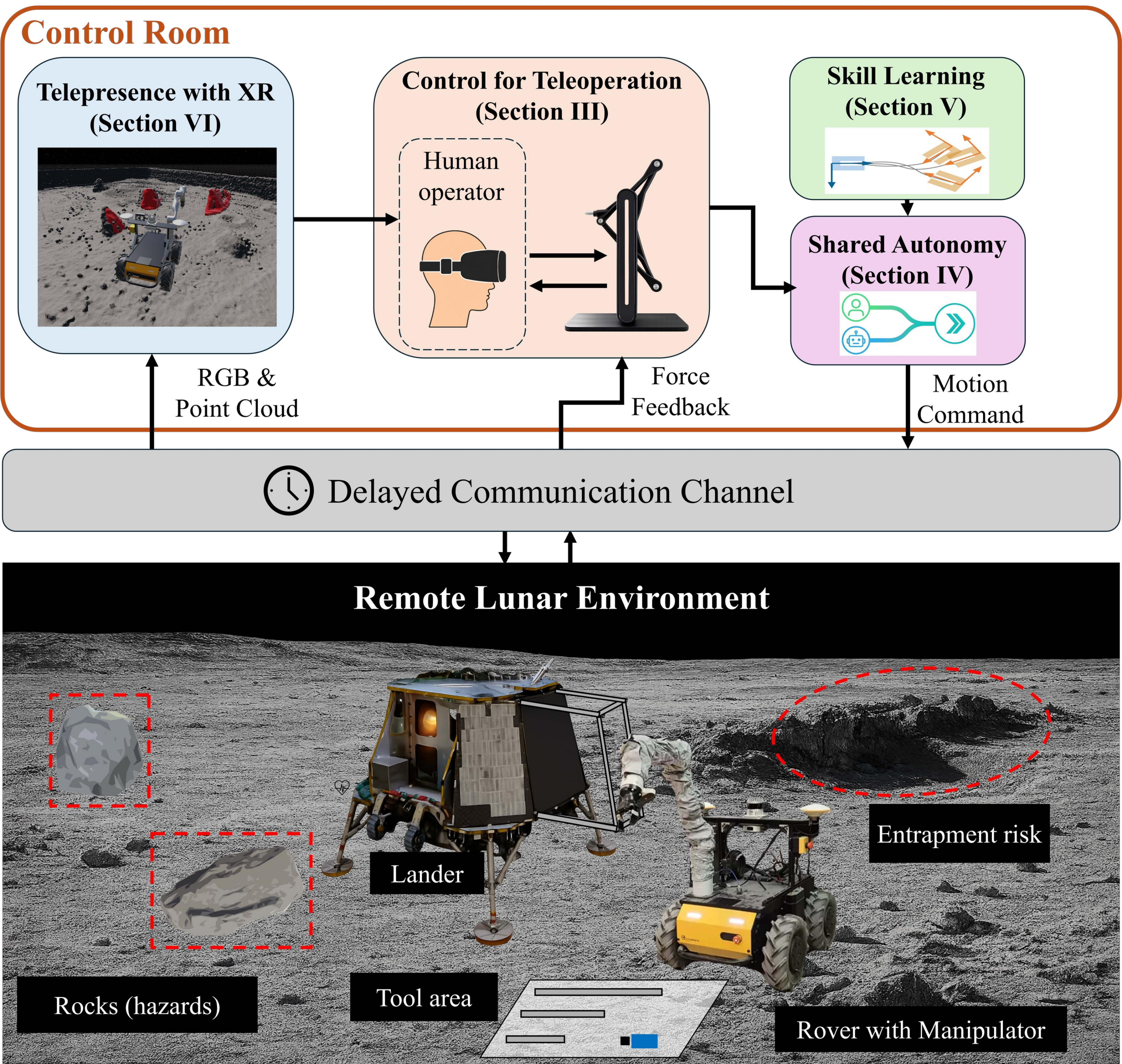}
    \caption{Overview of the presented framework. The architecture integrates teleoperation (Sec. \ref{sec:teleop}), shared autonomy (Sec. \ref{ssec:shared-autonomy}), skill learning (Sec. \ref{sec:skill-learning}), and XR-based telepresence (Sec. \ref{sec:XR}).}
    \label{fig:lunar_assembly_architecture}
    \vspace{-5mm}
\end{figure}

\huseyinrev{On the other hand}, for multi-second delays, autonomous task execution \huseyinrev{using Learning from Demonstration (LfD)} \huseyinrev{can be} preferred over direct teleoperation. \huseyinrev{LfD} enables robots to learn tasks from human demonstrations, such as kinesthetic teaching or teleoperation, without explicit programming. Demonstrated motion and force trajectories are encoded using machine learning models (e.g., Gaussian Mixture Models) and later reproduced autonomously \cite{calinon2016tutorial}. Extended Reality (XR) further mitigates delay by locally reconstructing the remote environment, improving situational awareness in perceptually challenging settings such as the LSP. This improves teleoperation through reduced workload, enhanced spatial awareness, and reliable obstacle detection over conventional 2D camera views \cite{frantz2025xrov}.

In this work, we present an Earth-to-Moon bilateral teleoperation system applied to an assembly task: extending a lunar landing module. An overview of the system architecture is illustrated in Fig. \ref{fig:lunar_assembly_architecture}. The task is first demonstrated through teleoperation in a delay-free \huseyinrev{lunar analog environment}, where the robot learns the required motion and force information using LfD. 
When this learned autonomy is deployed and fails, the operator takes over the task through direct teleoperation, assisted by shared control and additionally XR-based teleoperation. To keep this fallback usable despite Earth-Moon delay, we apply the Time Domain Passivity Approach (TDPA) to keep the communication channel passive, so the operator can reliably complete the task by direct control.
In this framework, a capability map of the robotic arm is integrated into the interface as visual guidance, assisting the operator in assessing and adjusting the robot’s position relative to the task target.
Finally, assuming a mostly static remote environment, we extend the shared control framework with XR-based teleoperation using a local reconstruction of the remote scene. This enables delay-free teleoperation in XR while rendering terrain hazards, such as rocks and deep pits, to help the operator safely avoid obstacles that could damage the rover. We present a range of teleoperation paradigms, spanning direct teleoperation, shared autonomy, learning from demonstration, and teleoperation with extended reality.


\section{System Description}
\subsection{Local Side}
The local side of the bilateral teleoperation system consists of a Haply Inverse3 haptic interaction device operated by the human user. Through the haptic device, the operator teleoperates the remote mobile robot and the 7-DoF Kinova Gen3 robot arm deployed in the lunar analog environment. The operator receives real-time visual feedback from the remote site through video streams transmitted over the communication network, enabling monitoring of the robot's state and its interaction with the environment. 
\subsection{Remote site: Analogue Facility}
\label{lunarfacility}
The remote end of the setup is hosted in an indoor lunar analogue facility \cite{LL}. This facility includes a 77sq.m. basalt test bed and controllable illumination, which together reproduce the low sun angles, long shadows and high-contrast lighting of the LSP. The mobile manipulator is a Clearpath Husky A300 carrying a Kinova Gen3 7-DoF arm with a Robotiq Hand-E gripper and a Bota six-axis force-torque sensor at the wrist. The perception stack includes an Ouster OS1 LiDAR and two Realsense d455 stereo cameras. The planned assembly scenario is a short traverse between a designated tool area and a lander mock-up on which a lateral frame extension is assembled, representative of early lunar surface-infrastructure assembly tasks.
\subsection{Communication setup}
\begin{figure}
    \centering
    \includegraphics[width=1.0\linewidth]{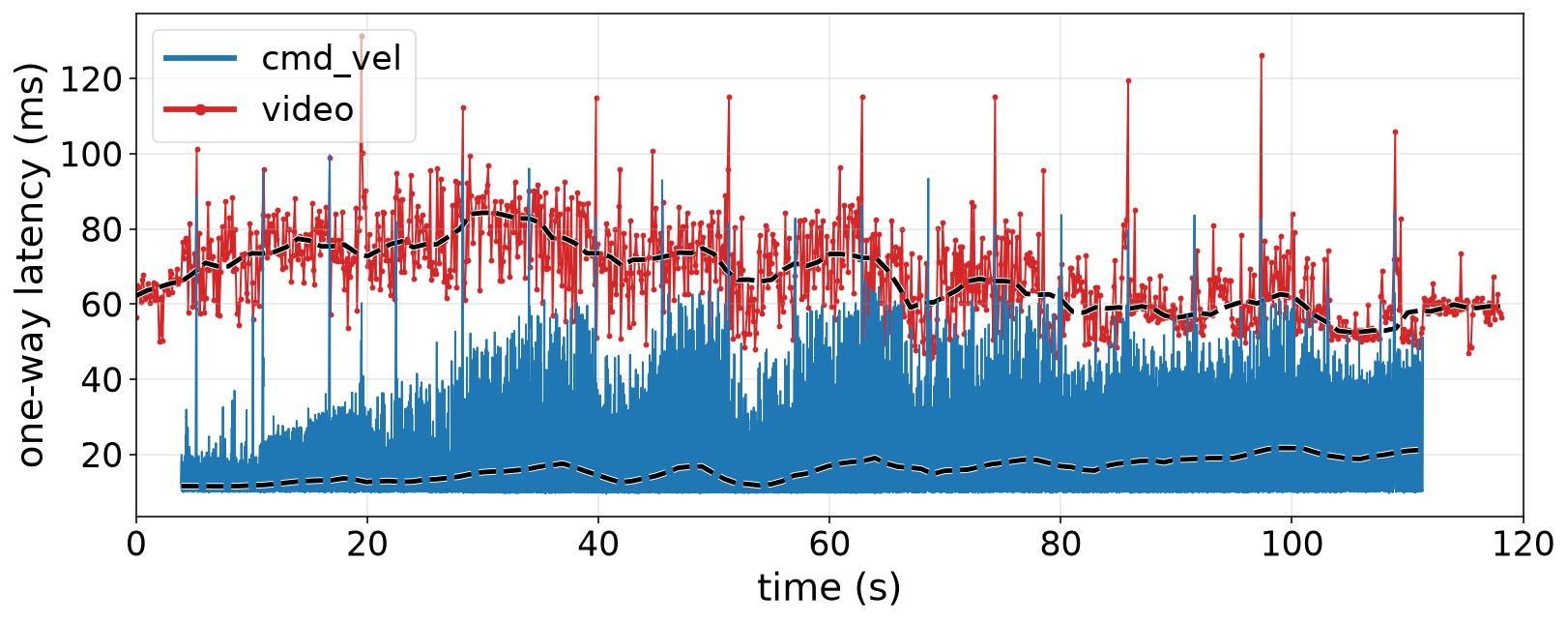}
    \caption{One-way latency during mobile-platform teleoperation with the Haply Inverse3 over a 2min drive. Commands are sent at 1kHz (blue), compressed video at 10Hz (red), with per-channel running medians (black, 5s window); one-way timing uses clock-synchronised endpoints. Command latency rises as the robot drives further from the wireless access point; spikes in the two channels frequently coincide, indicating shared network transients.}
    \label{fig:comms_latency}
    \vspace{-5mm}
\end{figure}

The teleoperation site and the remote site are separate institutions in two different countries, $\approx760$km apart, connected over the public Internet. ROS2 traffic is carried using the Zenoh middleware \cite{rmwzenoh} and routed through an authenticated relay on a public endpoint at the remote site. We validated the link by running teleoperation between the Husky in the analogue facility and a workstation in the separate institution through this relay, measuring a round-trip time of $\approx21$\,ms median ($\approx23$\,ms mean, $95\%\leq28$\,ms) and about $93$Mbit/s of throughput at the ROS2 level. Broken down per channel, the command path contributes $\approx$10\,ms one-way and the compressed video feed $\approx$68\,ms, both stable apart from rare sub-second excursions (Fig.~\ref{fig:comms_latency}). Packet loss is effectively zero for teleoperation commands and the camera feed at their operating rates (<100hz); it becomes non-zero only under haptic-rate command streams ($\approx$0.5-1\,kHz), remaining around $0.2\%$ in normal conditions and rising to $1.3$-$2.1\%$ only when such a stream coincides with a weak-signal position within the testbed, with no packet reordering or duplication observed. The native latency is therefore negligible against the multi-second delays of lunar operation, and the delay is instead imposed deliberately: a ROS2 node monitors the current RTT and injects additional delay to reach a controlled 3s round trip.
\section{Control for Time-delayed Bilateral Teleoperation} \label{sec:teleop}
A Cartesian impedance controller, for the remote robot, tracks the desired motion commanded by the haptic device. The controller that is explained in the Section~\ref{sec:cart_imp} achieves user defined impedance behavior with inertia shaping, gravity and Coriolis/centrifugal forces compensation. Disturbance observer is used to compensate the frictional and unmodeled effects in robot joint described in Section~\ref{sec:DOB}. To guarantee passivity under multi-second communication delays, the bilateral teleoperation system employs TDPA as detailed in Section~\ref{sec:TDPA}. 
\subsection{Cartesian Impedance Control with Force Tracking and Null Space Control}\label{sec:cart_imp}
Consider task-space dynamics of a redundant robot manipulator
\begin{equation} \label{system_model}
\boldsymbol{\Lambda}(\boldsymbol{q}) \ddot{\boldsymbol{x}}
+ \boldsymbol{\mu}_x(\boldsymbol{q},\dot{\boldsymbol{q}})\dot{\boldsymbol{x}} + \bar{\boldsymbol{J}}(\boldsymbol{q})^\top \boldsymbol{G}(\boldsymbol{q}) = \boldsymbol{f} + \boldsymbol{f}_{\mathrm{ext}},
\end{equation}
where $\bs{x} \in \mathbb{R}^m$ is the task-space coordinate, $\bs{\Lambda(q)} \in \mathbb{R}_{\succ 0}^{m \times m}$ is symmetric inertia matrix, $\bs{\mu}_x \bs{(q, \dot{q})}$ is Coriolis/centrifugal forces, $\bs{G(q)} \in \mathbb{R}^n$ is the gravitational forces, $\bs{J(q)} \in \mathbb{R}^{m \times n}$ is the Jacobian matrix and $\bar{\bs{J}}\bs{(q)} \in \mathbb{R}^{n \times m}$ is the weighted right pseudo inverse of $\bs{J(q)}$, $\bs{f} \in \mathbb{R}^{m}$ is the control signal, and $\bs{f}_\mathrm{ext} \in \mathbb{R}^m$ is the external forces. Given a reference task-space reference trajectory from the local device $\bs{x}_\mathrm{ref} = \bs{x}_{l, \mathrm{del}} \in  \mathbb{R}^m$, the error between current remote  pose and reference pose is $\Tilde{\bs{x}} := \bs{x} - \bs{x}_\mathrm{ref}$. We formulate the relationship between $\bs{f}_\mathrm{ext}$ and $\Tilde{\bs{x}}$ as
\begin{equation} \label{desired_impedance}
    \bs{\Lambda}_d\ddot{\tilde{\bs{x}}}(t) + \bs{\mu}_x(\bs{q,} \dot{\bs{q}})\dot{\tilde{\bs{x}}}(t) + \bs{D}_d\dot{\tilde{\bs{x}}}(t) + \bs{K}_d\tilde{\bs{x}}(t) = \bs{f}_\mathrm{ext},
\end{equation}
where $\bs{\Lambda}_d$, $\bs{D}_d$, $\bs{K}_d \in \mathbb{R}^{m \times m}$ are the positive definite diagonal desired (user-defined) inertia, damping, and stiffness matrices. The control force $\bs{F}$ that realizes~\eqref{desired_impedance} is calculated 
\begin{equation}
\label{control_signal}
\begin{aligned}
\bs{f} = {}&\bs{\Lambda}(\bs{q}) \ddot{\bs{x}}_d(t) 
  + \bigl(\bs{\Lambda}(\bs{q}) \bs{\Lambda}_d^{-1} - \bs{I}_m\bigr)\bs{f}_{\mathrm{ext}} + \bs{\mu}_x(\bs{q}, \dot{\bs{q}}) \dot{\bs{x}}(t) \\ 
  & + \bar{\bs{J}}(\bs{q})^\top \bs{G}(\bs{q}) - \bs{\Lambda}(\bs{q}) \bs{\Lambda}_d^{-1}
    \bigl(\bs{K}_d \tilde{\bs{x}}(t) + \bs{D}_d\dot{\tilde{\bs{x}}}(t) \bigr).
\end{aligned}
\end{equation}
Plugging~\eqref{control_signal} into~\eqref{system_model} results in the desired impedance behavior in~\eqref{desired_impedance}. Besides achieving the desired impedance behavior this control signal compensates the gravitational and Coriolis/centrifugal forces. This control signal can be written in the joint-space coordinates with the null space control torque as
\begin{equation}
    \bs{\tau}_\mathrm{ctrl} = \bs{J(q)}^\top\bs{F} + \bs{N(q)}\bs{\tau}_n + \boldsymbol{\tau}_f,
\end{equation}
where $\bs{\tau}_\mathrm{ctrl} \in \mathbb{R}^{n}$ is the joint torque that realizes the impedance in~\eqref{desired_impedance} and generates a null space torque with dynamically consistent null-space projector matrix $\bs{N(q)} := \bs{I}_n - \bs{J(q)}\bar{\bs{J}}\bs{(q)}$. The null space torque $\bs{\tau}_n$ is designed such that manipulator has a high manipulability index while avoiding singularities in task space. 
The wrench control term $\bs{\tau}_f \in \mathbb{R}^n$ facilitates tracking of a desired contact wrench $\bs{f}_d \in \mathbb{R}^m$ at the end effector during autonomous skill execution (Sec.~\ref{ssec:sequencing}).

\begin{figure*}[t]
    \centering
    \includegraphics[trim = 7mm 1mm 7mm 1mm,clip,width=0.9\linewidth]{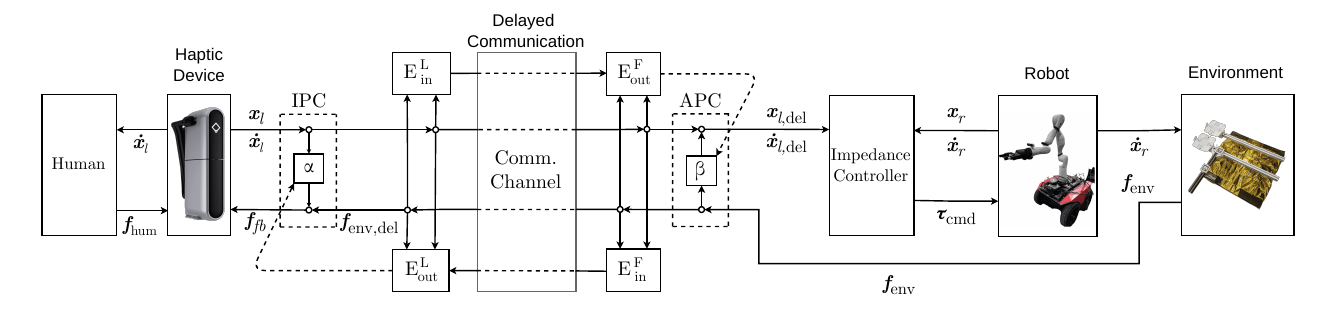}
    \caption{In bilateral teleoperation the haptic device's velocity is sent through the delayed communication channel, and environment force is returned to the operator as feedback $\boldsymbol{f}_{fb}$. The follower-side impedance controller uses the delayed leader velocity and remote robot velocity to compute $\boldsymbol{\tau}_{cmd}$. Energy flow is monitored at both ends of the communication channel ($E^\mathrm{L}_{\mathrm{in\&out}}, E^\mathrm{F}_{\mathrm{in\&out}}$) to detect passivity violations, triggering adaptive damping via an impedance-type (IPC, leader) or admittance-type (APC, follower) passivity controller.    
    }
    \label{fig:TDPA}
    \vspace{-5mm}
\end{figure*}
\subsection{Disturbance Observer}\label{sec:DOB}
Robot manipulators are generally suffer from joint friction, which degrades motion and torque tracking. To compensate for these effects, we employ the disturbance observer proposed in \cite{kim2019model}, which uses joint position, velocity, and torque measurements with a nominal plant model to estimate lumped disturbances, including friction and model uncertainties, and generate compensating torques.
We use the following motor model
\begin{equation}
\label{motor_model}
\boldsymbol{B}\bs{\ddot{\theta}} + \boldsymbol{\tau}_{\mathrm{meas}}
=
\boldsymbol{\tau}_{\mathrm{ctrl}} + \boldsymbol{\tau}_{d}.
\end{equation}
where $\bs{\theta} \in \mathbb{R}^{n}$ is the motor position,
$\boldsymbol{B} \in \mathbb{R}^{n\times n}$ is a diagonal matrix whose diagonal entries are the motor inertias,
$\boldsymbol{\tau}_{\mathrm{meas}}$ is the measured motor torque,
$\boldsymbol{\tau}_{\mathrm{ctrl}}$ is the commanded joint torque as obtained in Section~\ref{sec:cart_imp}, and
$\boldsymbol{\tau}_{d}$ denotes the joint friction and unmodeled uncertainties.
Using the measured motor torque $\bs{\tau}_{\mathrm{meas}}$ and the applied joint torque $\bs{\tau}_{\mathrm{ctrl}}$, the disturbance torque is estimated using the following PD-type observer:
\begin{equation}
    \hat{\bs{\tau}}_{d}
    =
    \bs{B}\bs{L}
    \left(
        \dot{\tilde{\bs{\theta}}}
        +
        \bs{L}_{p}\tilde{\bs{\theta}}
    \right),
\end{equation}
where $\bs{L}, \bs{L}_{p} \in \mathbb{R}^{n\times n}$ are diagonal positive-definite gain matrices, i.e.,
$\bs{L} \succ 0$ and $\bs{L}_{p} \succ 0$, and
$\tilde{\bs{\theta}}=\bs{\theta}-\bs{\theta}_{d}$ is the motor position error with respect to the desired motor position $\bs{\theta}_{d} = \bs{q}_d + \bs{K}_j^{-1}\bs{G}(\bs{q}_d)$, in which $\bs{q}_d$ is the desired link-side joint position, and $\bs{K}_j$ is the mechanical joint stiffness. Finally, the commanded joint torque is
\begin{equation}
    \bs{\tau}_\mathrm{cmd} = \bs{\tau}_\mathrm{ctrl} + \hat{\bs{\tau}}_{d}.
\end{equation}

\subsection{Ensuring Passivity with Time-Domain Passivity Approach}\label{sec:TDPA}
In time-delayed bilateral teleoperation passivity violations might occur because through time delay energy can be injected into the closed-loop system. This affects the interaction with the environment as it results in a buzzing or bouncing behavior of end effector. The TDPA has been extensively utilized to preserve passivity in force-feedback teleoperation systems with communication delays. It monitors the energy flow in real time using Passivity Observers (POs) and injects only the necessary amount of damping through Passivity Controllers (PCs) to ensure that the communication channel does not generate more energy than is supplied to it. Depending on the bilateral teleoperation architecture, impedance- or admittance-type PCs can be employed. 

We use a two-channel position--force measured architecture, shown in Fig. \ref{fig:TDPA}, in which position and velocity commands are transmitted to the remote side while environment interaction forces are returned as haptic feedback to the local side. In this architecture, the impedance-type PC (IPC) on the local side modifies the feedback force commanded to the haptic device, whereas the admittance-type PC (APC) on the remote side modifies the commanded velocity before the impedance controller. Depending on the activity level of the two communication ports, the corresponding PCs are activated to dissipate the excess energy generated by the time delay.

\section{Shared Autonomy and Operator Assistance}
In a remote and only partially known lunar environment, autonomous skill execution can fail due to degraded perception or situations outside the learned skill repertoire, while direct teleoperation places the whole task burden on the operator and becomes demanding under multi-second delay. We therefore keep the operator in the loop at two stages of the assembly task, combining the consistency of autonomous execution with human adaptability. A capability map assists the operator during base positioning (Sec.~\ref{ssec:capability}), and a shared autonomy mechanism lets the operator take over and correct the autonomous skill execution (Sec.~\ref{ssec:shared-autonomy}).

\subsection{Capability Map} \label{ssec:capability}

\begin{figure}[ht]
    \centering
    \includegraphics[width=1.0\linewidth]{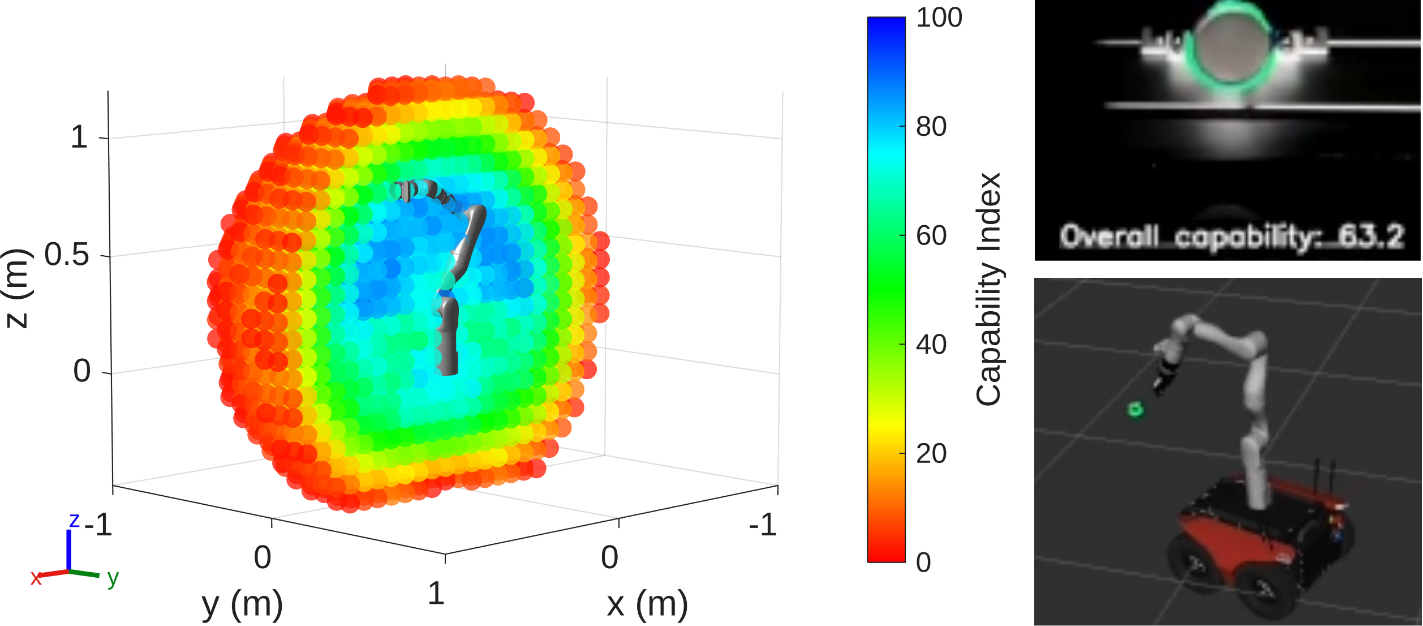}
    \caption{Left: Capability map of the robotic arm. Right: Live operator view showing object reachability and the relative robot--target position. The resolution and refresh rate depend on video transmission bandwidth.}
    \label{fig:capability_map}
    \vspace{-5mm}
\end{figure}

We incorporate a capability map \cite{ZACHARIAS13} to represent the operational suitability of the robotic arm throughout its workspace. The map is computed offline using the robot's kinematic model. Each discretized Cartesian position is assigned a capability score based on the percent of feasible end-effector poses satisfying the joint-limit constraints. As shown on the left of Fig.~\ref{fig:capability_map}, the scores are color-coded from red to blue, corresponding to low and high capability, respectively.

During execution, an eye-in-hand depth camera is used for object detection and three-dimensional reconstruction. The precomputed capability information is then overlaid onto both the two-dimensional camera image and the reconstructed point cloud, as shown on the right of Fig.~\ref{fig:capability_map}. As the mobile platform moves, the capability values associated with the detected object are updated online. This visualization allows the operator to determine whether the object lies within a favorable region of the manipulator workspace and to adjust the robot base pose accordingly.


\subsection{Shared Autonomy for Online Skill Correction}
\label{ssec:shared-autonomy}
A learned skill may require intervention during execution, for example due to unexpected environmental changes or sensor inaccuracies. We therefore let the operator take over control at any point of the autonomous skill execution (Sec.~\ref{sec:skill-learning}), by blending the end-effector twist of the autonomous motion ${\boldsymbol{v}}_{\mathrm{DS}} \in \mathbb{R}^m$ with the operator's input twist ${\boldsymbol{v}}_{l,\mathrm{del}} \in \mathbb{R}^m$: ${\boldsymbol{v}}_{d} = (1-\alpha_h)\,{\boldsymbol{v}}_{\mathrm{DS}} + \alpha_h\,{\boldsymbol{v}}_{l,\mathrm{del}}$. The operator's authority $\alpha_h\in[0,1]$ is obtained from the magnitude of ${\boldsymbol{v}}_{l,\mathrm{del}}$ through a sigmoid mapping and a recursive filter ensuring smooth transitions \cite{jadav_shared_2024}. The skill thus executes autonomously while the operator is passive, and authority shifts to the operator as soon as the input device is moved. Such interventions could indicate where the learned skill was inadequate, and could therefore be used to refine it through incremental learning \cite{heidersberger_lopal_2026}.

\begin{figure*}[t]
    \centering
    \includegraphics[trim={0 0 0 5mm},width=0.9\linewidth]{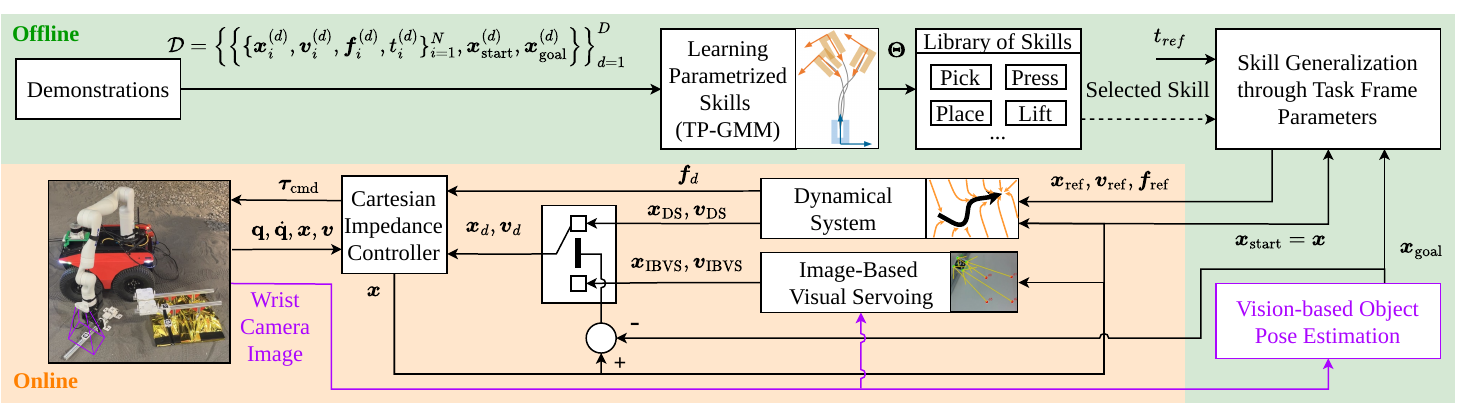}
    \caption{Overview of the framework for learning parametrizable skill primitives and their execution on the remote robot. The demonstration set $\mathcal{D}$ is encoded as a TP-GMM, parameterized by the start and goal frames ($\boldsymbol{x}_{\rm start}, \boldsymbol{x}_{\rm goal}$) for skill generalization in the current scene, and converted by an autonomous dynamical system into a disturbance-robust reference for the Cartesian impedance controller. Near the goal, image-based visual servoing corrects residual pose errors.}
    \label{fig:pipeline}
    \vspace{-5mm}
\end{figure*}

\section{Skill Learning for Remote Autonomous Assembly Tasks }
\label{sec:skill-learning}


Under delay, direct remote control can be challenging for the operator and lead to a high burden. Autonomous robotic skills can take over routine execution, allowing the operator to focus on monitoring and intervening when needed (Sec.~\ref{ssec:shared-autonomy}, ~\ref{ssec:preview}). We learn these skills from demonstrations using a task-parameterized representation, which allows to generalize to novel object configurations (Sec.~\ref{ssec:encoding}). Five demonstrations (containing $N$ samples of end-effector pose and wrench, and gripper state) were collected per skill using bilateral teleoperation (Sec.~\ref{sec:teleop}) in a delay-free lab environment, so that their quality was not degraded by the communication channel. 
We applied the learned skills on the remote robot to the assembly of a lateral frame extension (see Fig. \ref{fig:lunar_assembly_architecture}) through a behavior tree \cite{colledanchise2018behavior} sequencing of four skills: an aluminium rail is picked from the tool area, lifted clear of the ground, placed against the lander mock-up and pressed into two clip connectors for secure attachment.
Fig. \ref{fig:pipeline} gives an overview of the resulting pipeline, from skill learning to remote execution, which the remainder of this section describes in more detail.


\subsection{Learning Force-Aware, Task-Parameterized Skills}
\label{ssec:encoding}
The collected demonstrations are encoded as task-parameterized Gaussian Mixture Models (TP-GMM) with model parameters $\boldsymbol{\Theta}$~\cite{calinon2016tutorial}, forming a library of skills covering the behaviors required for the assembly task. 
Two properties of the TP-GMM encoding make it particularly suitable for the assembly skills. First, it jointly encodes the demonstrated motion and force profile, allowing to reproduce not only a path but also the required forces. Second, its task parameterization allows expressing each demonstration relative to a start and a goal frame ($\boldsymbol{x}_{\rm start}, \boldsymbol{x}_{\rm goal} \in SE(3)$); setting these task frames relative to object poses lets a skill generalize to new objects configurations without retraining.

\subsection{Autonomous Skill Execution on the Remote Robot} \label{ssec:sequencing}
Time-dependent movement primitives, such as TP-GMM, may generate excessive interaction forces when the end-effector is obstructed, as the commanded reference trajectory continues to advance in time. Therefore, the motion regressed via TP-GMR with input reference time $t_{\rm ref}$ is not commanded directly to the Cartesian impedance controller. Instead, an autonomous dynamical system (DS) \cite{jadav_shared_2024}, composed of state-based attraction and feedforward twist terms with respect to the regressed motion, generates the desired pose $\boldsymbol{x}_{\mathrm{DS}}$, twist ${\boldsymbol{v}}_{\mathrm{DS}}$, and wrench $\boldsymbol{f}_{ d}$.

Open-loop skill reproduction hinges on accurate object localization, which cannot always be assumed at the remote site. We used a wrist-mounted camera to localize objects via ArUco markers, but low resolution, calibration inaccuracies, and low-angle lighting caused positional errors of up to 1.6 cm.
Near the target, we therefore switch from DS-based control commands (${\boldsymbol{v}}_{\mathrm{DS}}$, $\boldsymbol{x}_{\mathrm{DS}}$) to image-based visual servoing (IBVS) commands (${\boldsymbol{v}}_{\mathrm{IBVS}}$, $\boldsymbol{x}_{\mathrm{IBVS}}$), using ArUco marker features as visual feedback to correct remaining misalignments and object pose changes in closed-loop during skill execution~\cite{chaumette2006visual}.

\subsection{Immersive Skill Preview and Reparameterization} \label{ssec:preview}
\begin{figure}
    \centering
    \includegraphics[trim={3mm 3mm 3mm 3mm}, width=0.5\linewidth]{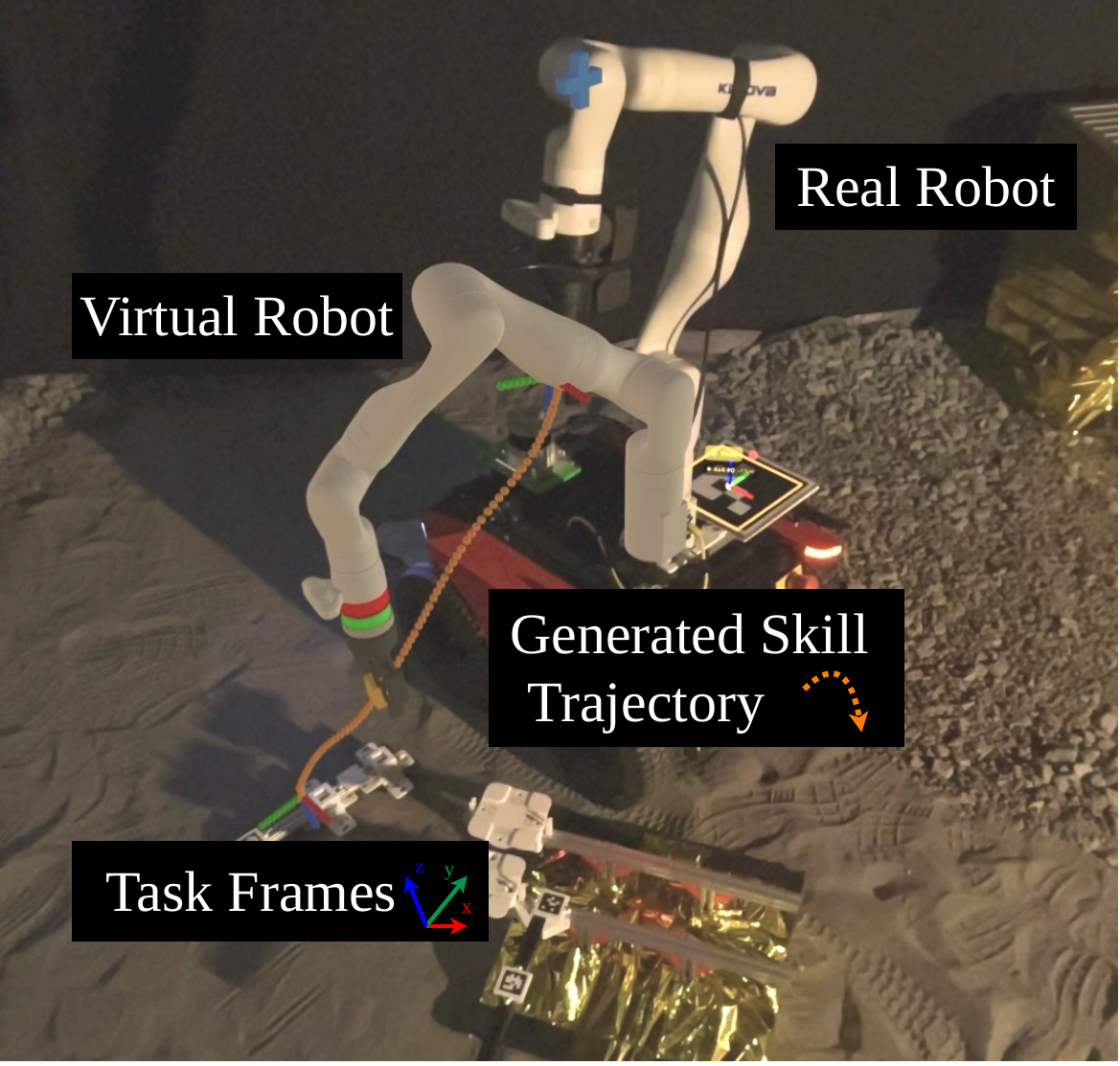}
    \caption{Immersive skill preview in extended reality. The visualized skill trajectory can be intuitively modified through interactive adjustment of its task frames and previewed via simulated execution before deployment on the robot.}
    \label{fig:skill-preview}
    \vspace{-5mm}
\end{figure}
Besides online corrections through shared autonomy (Sec.~\ref{ssec:shared-autonomy}), a skill can also be corrected before execution. In XR the operator can preview and intuitively modify the skill by manipulating its task parameters while maintaining awareness of the environment. The XR interface displays the skill's trajectory together with its task frames, rendered as objects that the user can grasp and reposition in 6-DoF (see Fig.~\ref{fig:skill-preview}). On release of these manipulatable task frames, the skill is regenerated and the updated trajectory is shown. An overlaid virtual robot can simulate the execution before deploying it on the real robot. 

\section{Telepresence through Extended Reality} 
\label{sec:XR}


Supervising assembly tasks under time delay requires the operator to anticipate distances and reachability, cues that conventional 2D video fails to convey. Perceiving hazards in the rover's vicinity is further complicated at the LSP, where low sun angles cast long shadows and produce a high dynamic range between lit and shadowed terrain, with parts of the scene falling into permanent or near-permanent shadow that inherently limits image-based perception. These conditions degrade the operator's ability to interpret the scene and introduce direct risks to safe assembly operations. To enhance operator awareness, we represent the remote scene as a 3D reconstruction viewed in XR, with hazards detected by a deep learning model and overlaid on it.

The remote scene is reconstructed through a Simultaneous Localization and Mapping (SLAM) framework \cite{RTABMap} running on the robot, which builds a 3D point-cloud map from LiDAR data. The 3D map is subsequently reconstructed as a 3D mesh and rendered in a Unity XR scene, giving the operator a navigable representation of the environment.

Hazards are detected on the RGB stream by a lightweight transformer segmentation model, a SegFormer \cite{SF} head on a MiT-b0 backbone, trained to segment rocks and craters. Each predicted mask is back-projected into a point cloud using the camera's depth channel, so that hazards appear as 3D geometry in the XR scene rather than as flat overlays. The segmentation model is trained on a lunar analogue dataset consisting of 3158 images, annotated rocks and craters with a model-assisted workflow in CVAT\cite{CVAT} with manual correction. The dataset is based on images from LunaPolaris \cite{LunaPolaris}, captured under low ($2.9^\circ$--$10.6^\circ$) and high ($12.3^\circ$--$38.7^\circ$) solar elevation angles.

\begin{figure}[t]
\centering 
{\includegraphics[width=0.80\linewidth]{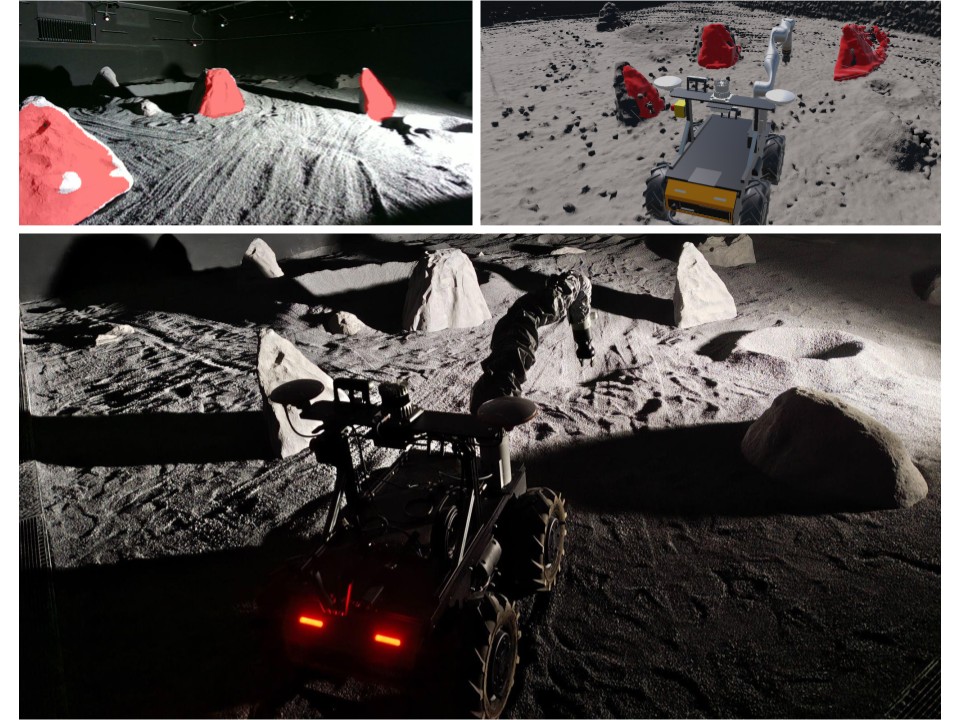}}
\caption{Bottom: Lunar analogue testbed with artificial rocks and two craters. Top-left: Nav-cam overlaid with segmented hazards. Top-right: XR scene with the segmented hazards as 3D overlays.}
\label{fig:vr-haz}
\vspace{-5mm}
\end{figure}

The full pipeline is integrated and tested in the lunar analogue environment of Sec.\ref{lunarfacility}, where detected hazards are visualized in the 3D reconstructed scene, as shown in Fig.~\ref{fig:vr-haz}. A residual domain gap from different camera hardware and parameters degrades the transfer, leaving several rocks and craters undetected. Segmentation inference takes $\approx10.3$ ms on a mobile GPU and is throttled to $5$ Hz.


We believe that using XR and 3D reconstruction techniques could potentially improve operator immersion and enable depth-aware teleoperation. Notably, using a 3D mesh to visualize the environment, including overlaying detected hazards as 3D information, enables operators to inspect the terrain from multiple viewpoints rather than 2D camera feeds with constrained views. However, quantifying the effectiveness of our XR framework on operator task performance and cognitive workload during teleoperation remains out of scope \johannesrev{of this work}.

\section{Conclusion}
In this work, the challenges of deploying bilateral teleoperation systems for a space application are addressed using different approaches, such as shared bilateral teleoperation, full autonomy with learning from demonstration, and remote environment reconstruction using extended reality.

Two limitations of the teleoperation setup were identified during experiments. First, during bilateral teleoperation the operator's situational awareness is constrained, since the only available visual feedback consisted of multiple camera views distributed across multiple windows, through which the operator must develop an understanding of the robot's surroundings and task state. This results in considerable cognitive load. Second, task completion degrades with an increase in communication delay. The operator was required to move at reduced velocity to avoid task failure or damage to the environment, as no knowledge of the final end-effector's position at the remote site is available.
Autonomous skill execution avoids the need for continuous teleoperation through the delayed channel. However, it remains sensitive to object localization errors and disturbances such as joint friction, motivating visual servoing and robust control. A possible extension could be the use of anomaly detection to abort the execution or request operator assistance when failures occur.


Future work will focus on separating operator intent from non-passive behavior to improve interaction transparency and integrating the interface with the reconstructed remote scene (Sec. \ref{sec:XR}) to enable skill preview and correction in remote settings. Additionally, we will focus on making the XR interface delay-aware, rendering the robot in its confirmed, commanded and predicted states, so the operator is not misled by a multi-second delay.
A human-subject evaluation will then compare the immersive interface against a 2D interface on the teleoperated assembly task involving navigation and manipulation, measuring task performance and operator workload under injected delay.

\begin{spacing}{0.97}
\bibliographystyle{IEEEtran}
\bibliography{Bib/refs}

@article{hokayem2006bilateral,
  author  = {Hokayem, Peter F. and Spong, Mark W.},
  title   = {Bilateral teleoperation: An historical survey},
  journal = {Automatica},
  volume  = {42},
  number  = {12},
  pages   = {2035--2057},
  year    = {2006}
}

@inproceedings{frantz2025xrov,
  title={XROV: An Immersive eXtended Reality (XR) Interface for Lunar Rover Mission Control},
  author={Frantz, Alexandre and Coloma, Sof{\'\i}a and Olivares-Mendez, Miguel A},
  booktitle={International Conference on Extended Reality},
  pages={42--61},
  year={2025},
  organization={Springer}
}

@inproceedings{kim2021impact,
  title={Impact of communication delay and temporal sensitivity on perceived workload and teleoperation performance},
  author={Kim, Eishi and others},
  booktitle={ACM symposium on applied perception 2021},
  pages={1--8},
  year={2021}
}

@article{dinc2024relaxing,
  title={Relaxing conservatism for enhanced impedance range and transparency in haptic interaction},
  author={Dinc, Huseyin Tugcan and Hulin, Thomas and Ott, Christian and Ryu, Jee-Hwan},
  journal={IEEE Transactions on Haptics},
  volume={17},
  number={1},
  pages={100--107},
  year={2024},
  publisher={IEEE}
}

@book{colledanchise2018behavior,
  title={Behavior trees in robotics and AI: An introduction},
  author={Colledanchise, Michele and {\"O}gren, Petter},
  year={2018},
  publisher={CRC Press}
}

@inproceedings{panzirsch2020safe,
  title={Safe interactions and kinesthetic feedback in high performance earth-to-moon teleoperation},
  author={Panzirsch, Michael and others},
  booktitle={2020 IEEE Aerospace Conference},
  pages={1--10},
  year={2020},
  organization={IEEE}
}

@article{calinon2016tutorial,
  author  = {Calinon, Sylvain},
  title   = {A tutorial on task-parameterized movement learning and retrieval},
  journal = {Intelligent Service Robotics},
  volume  = {9}, number = {1}, pages = {1--29}, year = {2016},
  doi     = {10.1007/s11370-015-0187-9}}

@article{heidersberger_lopal_2026,
	title = {{LOPAL}: {Local} {Performance}-{Aware} {Active} {Learning} {From} {Imperfect} {Demonstrations}},
	volume = {11},
	issn = {2377-3766},
	shorttitle = {{LOPAL}},
	doi = {10.1109/LRA.2026.3698364},
	number = {7},
	journal = {IEEE Robotics and Automation Letters},
	author = {Heidersberger, Johannes and Jadav, Shail and Lee, Dongheui},
	year = {2026},
	pages = {9040--9047},
}

@article{kim2019model,
  title={Model-free friction observers for flexible joint robots with torque measurements},
  author={Kim, Min Jun and Beck, Fabian and Ott, Christian and Albu-Sch{\"a}ffer, Alin},
  journal={IEEE Transactions on Robotics},
  volume={35},
  number={6},
  pages={1508--1515},
  year={2019},
  publisher={IEEE}
}

@inproceedings{jadav_shared_2024,
  author    = {Jadav, Shail and Heidersberger, Johannes and Ott, Christian and Lee, Dongheui},
  title     = {Shared Autonomy via Variable Impedance Control and Virtual Potential Fields for Encoding Human Demonstrations},
  booktitle = {2024 IEEE ICRA},
  pages     = {15151--15157}, year = {2024},
  doi       = {10.1109/ICRA57147.2024.10610761}}

@article{chaumette2006visual,
  author  = {Chaumette, Fran\c{c}ois and Hutchinson, Seth},
  title   = {Visual servo control. I. Basic approaches},
  journal = {IEEE Robotics \& Automation Magazine},
  volume  = {13}, number = {4}, pages = {82--90}, year = {2006},
  doi     = {10.1109/MRA.2006.250573}}

@inproceedings{LL,
	AUTHOR = {Ludivig, Philippe and others},
	TITLE = {Building a piece of the moon: Construction of two indoor lunar analogue environments},
	YEAR = {2020},
	BOOKTITLE = {71st International Astronautical Congress (IAC)}
}

@INPROCEEDINGS{LunaPolaris,
  author={van der Meer, Dave and Wong, Uland and Olivares-Mendez, Miguel},
  booktitle={2025 International Conference on Space Robotics (iSpaRo)}, 
  title={LunaPolaris: A stereo camera, point cloud and IMU dataset for future lunar exploration in polar regions}, 
  year={2025},
  doi={10.1109/iSpaRo66239.2025.11436512}}

@software{CVAT,
  title        = {Computer Vision Annotation Tool (CVAT)},
  author       = {{CVAT.ai Corporation}},
  year         = {2023},
  month        = {nov},
  day          = {6},
  version      = {2.25.0},
  doi          = {10.5281/zenodo.4009388},
  url          = {https://cvat.ai/},
  publisher    = {Zenodo},
  license      = {MIT},
  note         = {Repository: https://github.com/cvat}
}

@article{RTABMap,
  author  = {Labb\'{e}, Mathieu and Michaud, Fran\c{c}ois},
  title   = {{RTAB-Map} as an open-source lidar and visual simultaneous
             localization and mapping library for large-scale and
             long-term online operation},
  journal = {Journal of Field Robotics},
  volume  = {36},
  number  = {2},
  pages   = {416--446},
  year    = {2019},
  doi     = {10.1002/rob.21831}
}

@article{panzirsch2022exploring,
  title={Exploring planet geology through force-feedback telemanipulation from orbit},
  author={Panzirsch, Michael and others},
  journal={Science robotics},
  volume={7},
  number={65},
  pages={eabl6307},
  year={2022},
  publisher={American Association for the Advancement of Science}
}

@inproceedings{SF,
  title={SegFormer: Simple and Efficient Design for Semantic Segmentation with Transformers},
  author={Xie, Enze and Wang, Wenhai and Yu, Zhiding and Anandkumar, Anima and Alvarez, Jose M and Luo, Ping},
  booktitle={NeurIPS},
  year={2021}
}

@software{rmwzenoh,
  author       = {{Open Source Robotics Foundation}},
  title        = {{rmw\_zenoh}: {ROS 2} Middleware Interface for {Zenoh}},
  year         = {2024},
  url          = {https://github.com/ros2/rmw_zenoh},
  note         = {Accessed: 2026-08-06}
}

@article{ZACHARIAS13,
author = {Zacharias, Franziska and Borst, Christoph and Wolf, Sebastian and Hirzinger, Gerd},
title = {The Capability Map: A Tool to Analyze Robot Arm Workspaces},
journal = {International Journal of Humanoid Robotics},
volume = {10},
number = {04},
pages = {1350031},
year = {2013},
doi = {10.1142/S021984361350031X}}

@misc{fong2026viper,
  author       = {Fong, Terry},
  title        = {Volatiles Investigating Polar Exploration Rover ({VIPER})},
  howpublished = {Presentation, NASA Ames Research Center},
  month        = jan,
  year         = {2026},
  note         = {NASA Technical Reports Server (NTRS), Document ID 20260000802.}
}

@misc{nasa2026ignition,
  author       = {{NASA}},
  title        = {Ignition},
  howpublished = {\url{https://www.nasa.gov/ignition/}},
  year         = {2026},
  note         = {Accessed: 2026-08-07}
}
\end{spacing}

\end{document}